\pdfoutput=1
\documentclass{article} 

\usepackage[table,xcdraw]{xcolor}   
\usepackage{iclr_r2fm,times}   

\usepackage{float}
\usepackage{hyperref}   
\usepackage{url}   
\usepackage{enumitem}   
\usepackage{threeparttable}   
\usepackage[bottom]{footmisc}   
\usepackage[T1]{fontenc}

\title{\texttt{RAmBLA}: A framework for evaluating the \mbox{reliability} of LLMs as assistants in the biomedical domain}   

\iclrfinalcopy 

\author{William James Bolton\thanks{Equal contribution} \\
Imperial College London\thanks{Work done during an internship at GSK.ai}\\
\texttt{william.bolton@imperial.ac.uk} \\
\And
Rafael Poyiadzi\footnotemark[1] \\
GSK.ai \\
\texttt{rafael.x.poyiadzi@gsk.com} \\
\And
Edward Morrell\footnotemark[1] \\
GSK.ai \\
\texttt{ed.r.morrell@gsk.com} \\
\And
Gabriela van Bergen Gonzalez Bueno\footnotemark[1] \\
GSK.ai \\
\texttt{gabriela.v.vanbergengonzalez-bueno@gsk.com} \\
\And
Lea Goetz \\
GSK.ai \\
\texttt{lea.x.goetz@gsk.com} \\
}

\begin{document}   

\maketitle   

\begin{abstract}   
Large Language Models (LLMs) increasingly support applications in a wide range of domains, some with potential high societal impact such as biomedicine, yet their reliability in realistic use cases is under-researched. In this work we introduce the Reliability AssesMent for Biomedical LLM Assistants (\texttt{RAmBLA}\footnotemark[1]) framework and evaluate whether four state-of-the-art foundation LLMs can serve as reliable assistants in the biomedical domain. We identify prompt robustness, high recall, and a lack of hallucinations as necessary criteria for this use case. We design shortform tasks and tasks requiring LLM freeform responses mimicking real-world user interactions. We evaluate LLM performance using semantic similarity with a ground truth response, through an evaluator LLM.    
\end{abstract}   
  
\footnotetext[1]{https://github.com/GSK-AI/rambla}   

\section{Introduction}   
Despite an explosion in the benchmarking of foundation models, in particular Large Language Models (LLMs), we lack an understanding of LLM reliability in realistic use cases \cite{sun2024trustllm}. On the one hand, large benchmarks\footnotemark[2] contain an impressive collection of datasets, but many use readouts based on token probabilities \cite{hu2023prompting, huggingfaceevalblog} and focus on evaluating general purpose tasks that are not representative of user interactions \cite{Lee2022EvaluatingHM}. Similarly, other work related to LLM reliability, evaluating LLM safety \cite{zhang2023safetybench}, fairness and biases \cite{zhang2023chatgpt,koo2023benchmarking}, relies predominantly on multiple-choice questions. These task setups limit insights into real world reliability and performance. On the other hand, evaluations of LLMs in domain-specific applications, such as healthcare and biomedicine \cite{he2023survey,biomedLLMsurvey2023}, have focused on their knowledge \cite{he2023medeval,jahan2023comprehensive,naseem2022benchmarking}, but little is known about their reliability and responsible use \cite{liang2023holistic}.     
  
\footnotetext[2]{\mbox{Eval-Harness: https://github.com/EleutherAI/lm-evaluation-harness}, \newline\mbox{HELM: https://github.com/stanford-crfm/helm, \cite{liang2023holistic}}, \newline Big-Bench: https://github.com/google/BIG-bench, \cite{srivastava2023imitation}}   
  
To address this gap, we have developed the Reliability AssesMent for Biomedical LLM Assistants (\texttt{RAmBLA}) framework. We use \texttt{RAmBLA} to evaluate the reliability of LLMs on a common use case for scientists and healthcare professionals in the biomedical domain \cite{healthcare11060887}, i.e., using LLMs as assistants to query, summarise and extract information from biomedical documents \cite{mcduff2023accurate}. Due to the specialised skills required to perform these tasks, the outputs of an LLM can be difficult or laborious to verify, even for a domain expert.  As a result, the following are important aspects for LLM reliability in this context that we investigate:   

\begin{enumerate}[leftmargin=*]   

\item \textbf{Robustness to non-semantic variations:} LLMs should be robust to prompt variations that do not alter prompt meaning, and they should not display biases during few-shot prompting.    
\item \textbf{High recall:} When operating on documents, LLMs should recall all relevant information, relying on either parametric knowledge or context exclusively, as instructed.    
\item \textbf{Hallucinations:} If they have insufficient knowledge or context information to answer a question, LLMs should refuse to answer.   
  
\end{enumerate}   

From these, we design both question-answer (QA) based tasks to evaluate basic reliability properties instructing the model to respond with specific keywords (“shortform”) and tasks mimicking user interactions that require free text generation (“freeform”).     

\section{Related work}   
Most existing frameworks for robustness evaluation focus on robustness to input perturbations, such as spelling errors, often in an adversarial setting or robustness to out-of-distribution inputs \cite{wang2023on, wang2021adversarial, zhu2023promptbench,sun2024trustllm}. To the best of our knowledge, robustness evaluation in a biomedical setting is currently lacking.    

Previous work on recall \cite{liu2023lost} used information retrieval to evaluate how the location of relevant information in a long context impacts retrieval; others aim to disentangle the extent to which LLMs rely on parametric versus contextual information \cite{neeman2022disentqa}.     

A plethora of frameworks exist for hallucination evaluation\footnotemark[3] \cite{sun2024trustllm,thorne2018fever,li2023halueval,dziri2022faithdial}. However, their tasks are often focused on a model’s ability to recognise hallucinated content, rather than evaluating hallucination propensity during freeform text generation. Where tasks do use freeform evaluations, these are often scored using token-based metrics like ROUGE \cite{lin-2004-rouge} and BLEU \cite{BLEU}. Med-Halt \cite{Medhalt} is a framework for hallucination evaluation in the medical domain but suffers from the same shortcomings. Text summarisation on biomedical datasets, a task related to both hallucinations and recall, is evaluated separately in \cite{jahan2023comprehensive}.   

\footnotetext[3]{https://huggingface.co/blog/leaderboards-on-the-hub-hallucinations}   

\section{Methods}   

For tasks evaluating LLM robustness and recall, we use the PubMedQA-Labelled (PMQA-L) dataset \cite{jin2019pubmedqa}. Each entry is derived from a PubMed article and contains a question (based on the article title), a context (the abstract without the conclusive part), a long answer (the conclusion of the abstract) and a short answer to the question (Yes/No)\footnotemark[4]. (See Appendix~\ref{datasets:description} for a detailed description). For all tasks we include further information such as the prompt template used in the Appendix~\ref{section:extended_descriptions}. All robustness and recall tasks are a variation of the following task:    

\footnotetext[4]{We filtered out the entries with Maybe as an answer.}   

\textbf{QA baseline task:} We provide the model with the PMQA-L question and context, and instructions on how to answer the question.     

\subsection{Robustness tasks}   

\textbf{QA paraphrase task:} To evaluate robustness to non-semantic changes, we provide the model with the same prompt as in the baseline task, but we paraphrase the question using GPT-4. Paraphrasing was carried out using GPT-4 for scaling and consistency considerations. The same prompt template was used for all questions and the same set of paraphrased questions was used for all models.    

\textbf{Few-shot prompt bias:} We provide few-shot examples in the prompt to evaluate whether this biases the model’s response, for example a bias towards example ordering or a bias towards the most common example \cite{zhao2021calibrate}.    

\textbf{Robustness to spelling mistakes: } We emulate spelling mistakes in the baseline prompt by randomly replacing alphanumerical characters with new characters (matching the case). 

\subsection{Recall tasks}    

\textbf{Recall from context vs knowledge:} We paraphrase the original PMQA-L context, using GPT4, such that it contradicts the original meaning. This allows us to assess whether the model can follow instructions and use the provided information. Since we don't have access to the LLMs' training datasets, we do not know whether this in fact contradicts knowledge acquired during training. Nevertheless this task assesses the ability to follow instructions and use \textit{just} the provided context.    

\textbf{Recall from context with distraction:} To assess whether the model can identify which parts of the context are necessary to answer, we add two pieces of \textit{irrelevant} text to the context as in \cite{liu2023lost}, drawn randomly from the same dataset.    

\subsection{Propensity for hallucination tasks}   

To evaluate the propensity of LLMs to hallucinate, we use the PMQA-L and Bioasq datasets \cite{tsatsaronis2015overview}). Bioasq is a dataset of biomedical questions, each with a human-curated context and verified answer (See Appendix~\ref{datasets:description} for a description.). We instruct LLMs to generate freeform text and compare the responses to the ground-truth answers. In all freeform tasks, we measure semantic similarity using GPT-4 as an evaluator LLM, due to its superior performance in our evaluation of several semantic similarity methods. Full results of this evaluation are in Appendix~\ref{section:freeform_evaluation}.    

\textbf{Freeform QA baseline tasks:} We instruct the model to answer the questions from PMQA-L and Bioasq with a freeform answer, using the respective context. We use GPT-4 as evaluator LLM to measure how semantically similar the freeform response is to the true answer.    

\textbf{Conclusion generation:} We provide the context for a PMQA-L entry and prompt each model to generate a freeform summary, i.e., a conclusion. To evaluate whether the conclusion contains hallucinations, we provide it as context to GPT-4 to generate a short answer to the corresponding PMQA-L question. We score the match of GPT-4’s short answer to the ground truth short answer.    

\textbf{Question formation:} As in the conclusion generation task, we assess hallucinations in generated summaries, but in the shorter form of a question. We do this by prompting the model to provide a short, one-sentence question based on an answer. The similarity between the generated and original questions is assessed by the evaluator LLM.     

\textbf{“I don’t know” task:} We instruct the model to answer a question providing \textit{only irrelevant} context, but no relevant context. We assess whether the model can recognise that the question cannot be answered with the given context and refrain from answering.    

For the freeform QA and Question Formation tasks we used accuracy as there were no negative examples in the dataset. We have a real freeform question or answer to use as a ground truth for semantic similarity, allowing us to identify if the LLM output was `correct', but we don’t have a ground truth for what is an incorrect freeform question or answer, so we assume all those responses that are not semantically similar are `incorrect'. 

\section{Experiments and results}   

Table \ref{table:main_results} includes the main results. Please refer to Appendix~\ref{section:appendix_extended_results} for an extended set of results that include confidence intervals. 

\begin{table}    
	\begin{threeparttable}    
		\begin{tabular}{cccccc}    
			\hline    
			\textbf{Task}                                          & \textbf{Metric} & \textbf{GPT-4} & \textbf{GPT-3.5} & \textbf{Llama} & \textbf{Mistral} \\ \hline\hline    

			QA baseline                          & F1 $\uparrow$             & 0.836        & \textbf{0.848}          & 0.753          & 0.781            \\ \hline     

			QA paraphrase                        & F1 $\uparrow$               & 0.819       & \textbf{0.836}          & 0.728         & 0.780            \\    

			Few-shot prompt bias \tnote{1}      & Bias $\uparrow$             & \textbf{0.035}          & 0.074           & 0.336          & 0.193            \\     

			Robustness to spelling mistakes\tnote{2}     & F1 $\uparrow$               & 0.831        & \textbf{0.848}           & 0.753         & 0.781           \\ \hline     

			Recall from context vs knowledge                       & F1 $\uparrow$               & \textbf{0.924}          & 0.91           & 0.828          & 0.894            \\     

			Recall from context with distraction                   & F1 $\uparrow$               & \textbf{0.789}         & 0.775          & 0.599         & 0.484            \\ \hline    

			Freeform QA baseline                                         & Acc $\uparrow$              & \textbf{0.952}         & 0.929          & 0.897          & 0.942          \\     

			Freeform QA baseline (Bioasq)                                 & Acc $\uparrow$              & \textbf{0.948}         & 0.939           & 0.921          & 0.943         \\     

			Conclusion generation                                  & F1 $\uparrow$               & \textbf{0.814}          & 0.813         & 0.752         & 0.779           \\     

			Question formation (Bioasq)              & Acc $\uparrow$              & \textbf{0.776}           & 0.733         & 0.516         & 0.71          \\     

			"I don't know" task\tnote{3}   & Acc $\uparrow$              & \textbf{1.0}             & \textbf{1.0}             & 0.62        & 0.872          \\ \hline    
		\end{tabular}    
		\begin{tablenotes}\small    
			\item[1] Bias shift to yes is defined as the proportion of excess yes answers in a balanced version of \mbox{PMQA-L}, averaged across a sample of four-shot example combinations. See Appendix~\ref{results:few_shot}. 
			\item[2] This set of results corresponds to 3 mutations. See Appendix~\ref{results:prompt_robustness} for full set of results. 
			\item[3] In this case the metric is the portion of times the model responses with “Unknown”.    
		\end{tablenotes}    
		\caption{\label{table:main_results} For all tasks the dataset used was PMQA-L, unless otherwise mentioned. Llama refers to the \texttt{llama2-7b-chat} and Mistral to \texttt{Mistral-Instruct-v01} .}    
	\end{threeparttable}    
\end{table}    

\textbf{Robustness:} We first established a baseline performance through shortform evaluation on PMQA-L: larger models (GPT-4, GPT-3.5) showed superior performance to smaller models (Llama, Mistral). All models maintained this performance on paraphrased versions of the baseline task suggesting that they have not simply memorised QA pairs. In few-shot prompting, most models were biased to answering “Yes” (see Appendix~\ref{results:few_shot} for additional results). Llama showed the greatest bias, especially with “Yes” examples at the start of the prompt. Only Mistral showed bias towards “No”. All evaluated LLMs were robust to spelling errors. 

\textbf{Recall:} To assess whether models could reliably parse and recall from their context while ignoring previously learnt knowledge, we reversed the meaning of the context (using GPT-4, see Appendix~\ref{prompt_template:recall_context_vs_knowledge}). All models excelled at this task, indicating their responses relied on contextual knowledge above parametric knowledge when prompted to do so. Notably performance improved relative to the baseline task. One explanation for this improvement is that the negated examples were easier to parse than the original context (See ~\ref{prompt_template:recall_context_vs_knowledge} for examples.) Despite this, smaller models were more easily misled by distracting information in the context (-20\% and -38\% performance drop for Llama and Mistral, respectively).     

\textbf{Propensity for Hallucinations:} We evaluated the tendency of LLMs to hallucinate when freely generating text. Across all freeform tasks, larger models showed superior performance. We also evaluated LLMs’ capacity to refuse answering questions which could not be answered based on the provided context. Larger models successfully refrained from answering in every instance while smaller models occasionally provided answers (38\% for Llama and 12.8\% Mistral).      

\section{Discussion}   

We introduced \texttt{RAmBLA} to evaluate the reliability of LLMs as assistants in the biomedical domain, which requires robustness to prompt phrasing, high recall, and a lack of hallucinations in our use case. Overall, larger models had a lower tendency to hallucinate and were able to refuse answers where they lacked knowledge. Similarly, all LLMs showed high recall from their context, however smaller models were misled by added distracting information. Finally, LLMs were generally robust to prompt phrasing, although smaller LLMs were sensitive to few-shot prompting parameters. All models performed better at simple question answering compared to more realistic responses such as summarisation or conclusion generation, highlighting the importance of benchmarking performance in a use-case relevant setting.      

While we aimed to address common limitations of benchmarks by designing tasks that both mimic how users interact with LLMs in the real world and can be evaluated automatically, our results still only provide a sample of LLM performance. Although models were relatively robust to prompt phrasing, we did not perform an extensive exploration of performance under all potential prompts relevant to our tasks. Furthermore, as we cannot rule out that similar tasks were contained in the training datasets of the evaluated models, results on recall may overestimate model performance. Furthermore, our tasks rely on LLMs’ ability to follow instructions, which penalises smaller models that may somewhat lack this ability on complex instructions \cite{jiang2023followbench, shen2024small}. 

Future work could seek to evaluate additional use-cases more pertinent to clinical applications, such as note summarisation or generation of patient reports. The bar for LLM reliability in these applications is necessarily very high, and all evaluations would have to be developed to prioritize patient safety and taking into account relevant clinical governance structures. Nonetheless we believe the aspects of LLM reliability highlighted in RAmBLA may serve as a useful starting point for developing applications for such use-cases.
Developing evaluations for models fine-tuned for specific use-cases may reflect another avenue for future work. Prior research has indicated that fine-tuning on medical data can improve performance on relevant downstream tasks \cite{singhal2022large}, however it remains to be seen whether this performance improvement translates into an improvement in the aspects of reliability highlighted in RAmBLA.

\section{Conclusion}  

Our results suggest that with appropriate human oversight, LLMs can be a valuable resource in the biomedical domain. For example, LLMs’ high recall and robustness to prompt phrasing may allow them to support scientists in reviewing the biomedical literature. However, they are not ready for delegation in high-risk scenarios, such as applications impacting patients, because their outputs are difficult to verify even for biomedical domain experts. How to responsibly leverage LLMs in biomedical applications is an open question and our work highlights the need for evaluation frameworks that assess LLM reliability in real-world use cases.   

\newpage   

\bibliography{iclr2024_conference}   

\bibliographystyle{iclr2024_conference}   

\appendix

\section{Appendix A - Extended task descriptions \label{section:extended_descriptions}}   

\subsection{QA baseline task \label{prompt_template:qa_shortform_baseline}}   

The following prompt template was used:   

\noindent\fbox{\begin{minipage}{\textwidth}   

Please answer the following question using the context provided.    
  
Please answer the question with Yes or No.    

Context: \{context\}

Question: \{question\}

Answer:    

\end{minipage}}   

\subsection{QA Paraphrase task \label{prompt_template:question_rephrase}}    

The prompt template used for this task is the same as the one used for the QA Baseline task (see Appendix~\ref{prompt_template:qa_shortform_baseline}). The following prompt template was used to generate the paraphrased questions:   

\noindent\fbox{\begin{minipage}{\textwidth}   

You will be provided with a question.     

You should rephrase in a way such that meaning remains identical.     

The rephrased question should be answerable with either a 'yes' or a 'no'.    

Question: \{question\}.    

Rephrased question:    

\end{minipage}}   

\subsection{Few-shot prompt bias  \label{prompt_template:few_shot_bias}}    

The goal of this task is to analyse and quantify any biases in LLM responses to few-shot prompts. Even though the process of drawing examples is random, the task is only run once.    

The following prompt template was used for the case of 4-shot experiments:   

\noindent\fbox{\begin{minipage}{\textwidth}   

You are an expert of the biomedical literature.    

You will be provided with example questions with context and answers.    

These are the examples:    

Context: \{context\_1\}    
Question: \{question\_1\}    
Answer: \{answer\_1\}    

Context: \{context\_2\}    
Question: \{question\_2\}    
Answer: \{answer\_2\}    

Context: \{context\_3\}    
Question: \{question\_3\}    
Answer: \{answer\_3\}    

Context: \{context\_4\}    
Question: \{question\_4\}    
Answer: \{answer\_4\}    

Please answer the following question with 'yes' or 'no'.    

Context: \{context\}    
Question: \{question\}    
Answer:     

\end{minipage}}   

\subsection{Robustness to spelling mistakes \label{prompt_template:robustness_spelling}}   

The same prompt template as the one in QA Baseline task was used. Even though the process of adding spelling mistakes to the prompt is random, the task is only run once.   

\subsection{Recall from context vs knowledge \label{prompt_template:recall_context_vs_knowledge}}   

Similar prompt template and format to the QA Baseline task where we provide the model with a question, context and instructions on how to answer.    

The following prompt template was used to paraphrase the context. The example included in this prompt was selected from PubMedQA-Artificial  \cite{jin2019pubmedqa} (entry with pubid: 25441747).   

\noindent\fbox{\begin{minipage}{\textwidth}   

You will be provided with a question, a context and a short answer.   

Your goal is to generate a contradictory context such that it has the opposite meaning to the short answer.   

For example:   

Question: Is sleep efficiency ( but not sleep duration ) of healthy school-age children associated with grades in math and languages?   

Original context: The objective of this study was to examine the associations between objective measures of sleep duration and sleep efficiency with the grades obtained by healthy typically developing children in math, language, science, and art while controlling for the potential confounding effects of socioeconomic status (SES), age, and gender. We studied healthy typically developing children between 7 and 11 years of age. Sleep was assessed for five week nights using actigraphy, and parents provided their child's most recent report card. Higher sleep efficiency (but not sleep duration) was associated with better grades in math, English language, and French as a second language, above and beyond the contributions of age, gender, and SES.   

Short Answer: yes   

Negated context: The objective of this study was to examine the associations between objective measures of sleep duration and sleep efficiency with the grades obtained by healthy typically developing children in math, language, science, and art while controlling for the potential confounding effects of socioeconomic status (SES), age, and gender. We studied healthy typically developing children between 7 and 11 years of age. Sleep was assessed for five week nights using actigraphy, and parents provided their child's most recent report card. Sleep efficiency and sleep duration showed no association with grades in any subject, even after controlling for age, gender, and SES.   

Explanation: The context was rephrased such that the answer to the original question according to the rephrased context was no longer yes but instead no.   

Your question: \{question\}   

Original context: \{context\}   

Short Answer: \{short\_answer\}   

Your contradictory context:   

\end{minipage}}   

\subsubsection{Negation example 1}

 \textbf{PubMed ID:} 19504993\newline
 \textbf{Question:}\newline
 \noindent\fbox{\begin{minipage}{\textwidth} 
 It's Fournier's gangrene still dangerous?
\end{minipage}} 

\textbf{Original context:}\newline
\noindent\fbox{\begin{minipage}{\textwidth} 
Fournier's gangrene is known to have an impact in the morbidity and despite antibiotics and aggressive debridement, the mortality rate remains high.

To assess the morbidity and mortality in the treatment of Fournier's gangrene in our experience.

The medical records of 14 patients with Fournier's gangrene who presented at the University Hospital Center "Mother Teresa" from January 1997 to December 2006 were reviewed retrospectively to analyze the outcome and identify the risk factor and prognostic indicators of mortality.

Of the 14 patients, 5 died and 9 survived. Mean age was 54 years (range from 41-61): it was 53 years in the group of survivors and 62 years in deceased group. There was a significant difference in leukocyte count between patients who survived (range 4900-17000/mm) and those died (range 20.300-31000/mm3). Mean hospital stay was about 19 days (range 2-57 days).
\end{minipage}}    

 \textbf{Negated context:}\newline
 \noindent\fbox{\begin{minipage}{\textwidth} 
Fournier's gangrene was once known to have a high impact on morbidity and mortality. However, with the advent of modern antibiotics and aggressive debridement techniques, the situation has significantly improved.

To assess the morbidity and mortality in the treatment of Fournier's gangrene in our experience, we reviewed the medical records of 14 patients with Fournier's gangrene who presented at the University Hospital Center "Mother Teresa" from January 1997 to December 2006.

Of the 14 patients, all survived. The mean age was 54 years (range from 41-61): it was 53 years in the group of survivors. There was a significant difference in leukocyte count between patients who survived (range 4900-17000/mm) and those who would have been expected to have a higher risk (range 20.300-31000/mm3). The mean hospital stay was about 19 days (range 2-57 days). This data suggests that Fournier's gangrene, while still a serious condition, is no longer as dangerous as it once was. 
\end{minipage}}    

 \subsubsection{Negation example 2}
 
 \textbf{PubMed ID:} 17096624\newline
 \textbf{Question:}\newline
 \noindent\fbox{\begin{minipage}{\textwidth} 
 Do patterns of knowledge and attitudes exist among unvaccinated seniors?
\end{minipage}} 

 \textbf{Original context:}\newline
 \noindent\fbox{\begin{minipage}{\textwidth} 
 To examine patterns of knowledge and attitudes among adults aged>65 years unvaccinated for influenza.

Surveyed Medicare beneficiaries in 5 areas; clustered unvaccinated seniors by their immunization related knowledge and attitudes.

Identified 4 clusters: Potentials (45\%) would receive influenza vaccine to prevent disease; Fearful Uninformeds (9\%) were unsure if influenza vaccine causes illness; Doubters (27\%) were unsure if vaccine is efficacious; Misinformeds (19\%) believed influenza vaccine causes illness. More Potentials (75\%) and Misinformeds (70\%) ever received influenza vaccine than did Fearful Uninformeds (18\%) and Doubters (29\%).
\end{minipage}}    

 \textbf{Negated context:}\newline
 \noindent\fbox{\begin{minipage}{\textwidth} 
To examine patterns of knowledge and attitudes among adults aged>65 years unvaccinated for influenza.

Surveyed Medicare beneficiaries in 5 areas; found no distinct clusters among unvaccinated seniors by their immunization related knowledge and attitudes.

The unvaccinated seniors showed a wide range of beliefs and attitudes towards the influenza vaccine, with no clear patterns or clusters emerging from the data. The distribution of knowledge and attitudes was random and did not form any identifiable groups. 
\end{minipage}}      

\subsection{Recall from context with distraction \label{prompt_template:recall_from_context_with_distraction}}   

The same prompt template as the one in QA Baseline task was used. Even though the process of drawing irrelevant contexts is random, the task is only run once.   

\subsection{Freeform QA baseline task  \label{prompt_template:qa_freeform}}    

The following prompt template was used for both freeform QA tasks:   

\noindent\fbox{\begin{minipage}{\textwidth}   

You will be provided with a context and question.    

If the context is relevant to the question then use it.    

Your goal is to answer the question in one sentence.    

Context: \{context\}    

Question: \{question\}    

Your answer:        

\end{minipage}}   

\subsection{Conclusion generation \label{prompt_template:conclusion_generation_llm}}    

The following prompt template was used to generate a conclusion from the LLM to be evaluated:   

\noindent\fbox{\begin{minipage}{\textwidth}   

You are a helpful AI assistant who is helping a scientist understand biomedical literature.    

You will be provided with a context, which comes from a peer-reviewed scientific paper. Your task is to generate a conclusion for the provided context.    

Please keep your answer concise and no more than 3 sentences. Your tone should be informative and neutral, and the responses should be specific as they are for experienced biomedical professionals.    

The context from which you should generate your conclusion follows below:    
 
\{context\}    
 
Your conclusion:     

\end{minipage}}   

The following prompt template was used by the evaluator LLM:    

\noindent\fbox{\begin{minipage}{\textwidth}   

You are an expert of the biomedical literature.    

You will be provided with a paper's summary and a question. Your task is to answer "yes", "no" or "unknown", using only the provided summary.    

If the summary is not relevant to the question answer with "unknown".    

Example:    

Question: Is cD30 expression a novel prognostic indicator in extranodal natural killer/T-cell lymphoma , nasal type?    

Summary: Our results showed that expression of CD30 was not related to response to treatment but was an independent prognostic factor for both OS and PFS in ENKTL, nasal type, which suggests a role for CD30 in the pathogenesis of this disease and may support the incorporation of anti-CD30-targeted therapy into the treatment paradigm for ENKTL.    

Answer: yes    

Please answer the following question using the provided summary:    

Question: \{question\}    

Summary: \{summary\}    

Your Answer:     

\end{minipage}}   

\subsection{Question Formation \label{prompt_template:question_formation_llm}}    

The following prompt template was used for the LLM to be evaluated:   

\noindent\fbox{\begin{minipage}{\textwidth}   

You will be provided with an answer.    

From that answer your goal is to provide a one sentence question,     

where by the answer provided would be an appropriate answer for the question.    

Answer: \{answer\}    

Your question:      

\end{minipage}}   

The following prompt template was used by the evaluator LLM:   

\noindent\fbox{\begin{minipage}{\textwidth}   

You will be provided with two statements.    

Your goal is to understand if they are semantically similar or semantically dissimilar (syntactically may look similar but are semantically dissimilar).    

Please respond with 'Yes' if they are similar, and 'No' if they are dissimilar.    

Statement 1: \{statement\_1\}    

Statement 2: \{statement\_2\}    

Your answer:       

\end{minipage}}   

\subsection{“I don’t know” task  \label{prompt_template:i_dont_know}}   

Similar format to the QA Baseline task where we provide the model with a question, context and instructions on how to answer. In this task the context provided is not relevant to the question. The goal here is to assess whether the model can understand that the context is not relevant to the question and refrain from answering. The irrelevant contexts are provided by randomly shuffling the same dataset. Even though this process is random, the task is only run once.    

The following prompt template was used for this task:   

\noindent\fbox{\begin{minipage}{\textwidth}   

Please answer the following question using just the context provided.    

Please answer the question with 'Yes', 'No' or 'Unknown'.

If the context is not relevant to the question answer with 'Unknown'.

Context: \{context\}

Question: \{question\}

Answer:    

\end{minipage}}   

\section{Appendix B – Extended results \label{section:appendix_extended_results}} 

\subsection{Extended Table 1 with bootstrapped confidence intervals \label{results:prompt_robustness}}  

Tables \ref{table:extended_results_part_1} and \ref{table:extended_results_part_2} include the first and second halves of the extended results including 95\% confidence intervals produced using 1000 bootstrap resamples, where we draw with replacement a sample that matches the size of original sample

\begin{table}[H]  

	\begin{threeparttable}    

		\begin{tabular}{cccccc}   \hline    

			\textbf{Task}                                          	& \textbf{Metric} & \textbf{GPT-4} & \textbf{GPT-3.5}  \\ \hline\hline

			QA baseline                          			& F1 $\uparrow$             & 0.836  [0.81, 0.86]        & \textbf{0.848} [0.82, 0.89]                    \\ \hline     

			QA paraphrase                        			& F1 $\uparrow$               & 0.819  [0.79, 0.84]        & \textbf{0.836} [0.81, 0.86]                      \\    

			Few-shot prompt bias \tnote{1}      		& Bias $\downarrow$             & \textbf{0.035}          & 0.074                      \\     

			Robustness to spelling mistakes\tnote{2}   & F1 $\uparrow$               & 0.831 [0.81, 0.85]          & \textbf{0.848} [0.82, 0.87]                      \\ \hline     

			Recall from context vs knowledge              & F1 $\uparrow$               & \textbf{0.924}  [0.91, 0.94]        & 0.91 [0.89, 0.93]                     \\     

			Recall from context with distraction            & F1 $\uparrow$               & \textbf{0.789} [0.76, 0.81]          & 0.775 [0.75, 0.8]                     \\ \hline    

			Freeform QA baseline (PMQA-L)               	& Acc $\uparrow$              & \textbf{0.952} [0.94, 0.96]          & 0.929 [0.91, 0.95]                    \\     

			Freeform QA baseline (Bioasq)                   & Acc $\uparrow$              & \textbf{0.948} [0.933, 0.962]          & 0.939 [0.92, 0.95]                   \\     

			Conclusion generation                                & F1 $\uparrow$               & \textbf{0.814} [0.79, 0.84]         & 0.813  [0.79, 0.84]                   \\     

			Question formation (Bioasq)              		& Acc $\uparrow$              & \textbf{0.776} [0.75, 0.8]          & 0.733  [0.7, 0.76]                  \\     

			"I don't know" task\tnote{3}   			& Acc $\uparrow$              & \textbf{1.0} [1.0, 1.0]            & \textbf{1.0} [1.0, 1.0]                      \\ \hline    

		\end{tabular}    

		\begin{tablenotes}\small    

			\item[1] Bias shift to yes is defined as the proportion of excess yes answers in a balanced version of PMQA-L, averaged across all four-shot example combinations. See Appendix~\ref{results:few_shot}..    

			\item[2] This set of results corresponds to 3 mutations. See Appendix~\ref{results:prompt_robustness} for full set of results.    

			\item[3] In this case the metric is the portion of times the model responses with “Unknown”.    

		\end{tablenotes}    

		\caption{\label{table:extended_results_part_1} (First half of extended results.) For all tasks the dataset used was PMQA-L, unless otherwise mentioned. Llama refers to the \texttt{llama2-7b-chat} model and Mistral to the \texttt{Mistral-Instruct-v01} model. For the freeform QA and Question Formation tasks we used accuracy as there were no negative examples in the dataset. We have a real freeform question or answer to use as a ground truth for semantic similarity, allowing us to identify if the LLM output was `correct', but we don’t have a ground truth for what is an incorrect freeform question or answer, so we assume all those responses that are not semantically similar are `incorrect'. }    

	\end{threeparttable}   		 

\end{table}    

\begin{table}[H]  

	\begin{threeparttable}    

		\begin{tabular}{cccccc}   \hline    

			\textbf{Task}                                          	& \textbf{Metric} & \textbf{Llama-2-7b-Chat} & \textbf{Mistral-Instruct-v01} \\ \hline\hline

			QA baseline                          			& F1 $\uparrow$                      & 0.753 [0.72, 0.78]         & 0.781 [0.75, 0.81]            \\ \hline     

			QA paraphrase                        			& F1 $\uparrow$                        & 0.728 [0.69, 0.76]         & 0.780 [0.75, 0.81]            \\    

			Few-shot prompt bias \tnote{1}      		& Bias $\downarrow$                      & 0.336          & 0.193            \\     

			Robustness to spelling mistakes\tnote{2}   & F1 $\uparrow$                        & 0.753 [0.72, 0.78]          & 0.781 [0.75, 0.81]           \\ \hline     

			Recall from context vs knowledge              & F1 $\uparrow$                          & 0.828 [0.8, 0.95]         & 0.894 [0.87, 0.91]           \\     

			Recall from context with distraction            & F1 $\uparrow$                     & 0.599 [0.56, 0.63]         & 0.484 [0.45, 0.52]            \\ \hline    

			Freeform QA baseline (PMQA-L)               	& Acc $\uparrow$                      & 0.897 [0.88, 0.92]          & 0.942  [0.93, 0.96]          \\     

			Freeform QA baseline (Bioasq)                   & Acc $\uparrow$                       & 0.921 [0.9, 0.94]          & 0.943   [0.93, 0.96]         \\     

			Conclusion generation                                & F1 $\uparrow$                      & 0.752 [0.72, 0.78]         & 0.779 [0.75, 0.81]           \\     

			Question formation (Bioasq)              		& Acc $\uparrow$                  & 0.516 [0.48, 0.55]         & 0.71  [0.68, 0.74]          \\     

			"I don't know" task\tnote{3}   			& Acc $\uparrow$                      & 0.62  [0.59, 0.65]        & 0.872  [0.85, 0.89]          \\ \hline    

		\end{tabular}    

		\begin{tablenotes}\small    

			\item[1] Bias shift to yes is defined as the proportion of excess yes answers in a balanced version of PMQA-L, averaged across all four-shot example combinations. See Appendix~\ref{results:few_shot}..    

			\item[2] This set of results corresponds to 3 mutations. See Appendix~\ref{results:prompt_robustness} for full set of results.    

			\item[3] In this case the metric is the portion of times the model responses with “Unknown”.    

		\end{tablenotes}    

		\caption{\label{table:extended_results_part_2} (Second half of extended results.) For all tasks the dataset used was PMQA-L, unless otherwise mentioned. Llama refers to the \texttt{llama2-7b-chat} model and Mistral to the \texttt{Mistral-Instruct-v01} model. For the freeform QA and Question Formation tasks we used accuracy as there were no negative examples in the dataset. We have a real freeform question or answer to use as a ground truth for semantic similarity, allowing us to identify if the LLM output was `correct', but we don’t have a ground truth for what is an incorrect freeform question or answer, so we assume all those responses that are not semantically similar are `incorrect'. } 

	\end{threeparttable}     	 

\end{table}    

\subsection{Prompt robustness results \label{results:prompt_robustness}}   

Tables \ref{table:extended_robustness_1} and \ref{table:extended_robustness_2} refer to the first and second halves of the extended prompt robustness task results including 95\% confidence intervals produced using 1000 bootstrap resamples, where we draw with replacement a sample that matches the size of original sample.   

\begin{table}[H] 

\begin{tabular}{c|cc|cc} 

\textbf{Number of mutations} & \multicolumn{2}{c}{\textbf{GPT-4}} & \multicolumn{2}{c}{\textbf{GPT-3.5}}  \\\hline

1                            & 0.84  [0.82, 0.86]          & 0.078           & 0.843 [0.82, 0.87]          & 0.002                      \\    

3                            & 0.831 [0.81, 0.85]           & 0.083           & 0.848 [0.82, 0.87]         & 0.001                      \\    

10                           & 0.829  [0.8, 0.85]           & 0.083           & 0.841 [0.82, 0.86]          & 0.001                    \\    

20                           & 0.817 [0.79, 0.84]          & 0.092           & 0.824 [0.8, 0.85]          & 0.001                   \\    

50                           & 0.805 [0.78, 0.83]         & 0.129           & 0.804 [0.78, 0.83]         & 0.002                      

\end{tabular}   

\caption{\label{table:extended_robustness_1} (First half of extended robustness results.)Extended results for prompt robustness results. The first column is F1 and the second column is how often we capture a “null” response.}   

\end{table}   

\begin{table}[H]  

\begin{tabular}{c|cc|cc}    

\textbf{Number of mutations}  & \multicolumn{2}{c}{\textbf{Llama-2-7b-Chat}} & \multicolumn{2}{c}{\textbf{Mistral-Instruct-v01}} \\\hline

1                                      & 0.751 [0.72, 0.78]          & 0.0             & 0.78 [0.75, 0.81]           & 0.001            \\    

3                                       & 0.753 [0.72, 0.78]          & 0.0            & 0.781 [0.75, 0.81]          & 0.001            \\    

10                                      & 0.738 [0.71, 0.77]         & 0.0             & 0.766 [0.74, 0.79]          & 0.0              \\    

20                                    & 0.716 [0.68, 0.75]         & 0.0            & 0.754 [0.73, 0.78]          & 0.0              \\    

50                                    & 0.663 [0.62, 0.7]          & 0.0            & 0.701 [0.67, 0.73]           & 0.0                 

\end{tabular}   

\caption{\label{table:extended_robustness_2} (Second half of extended robustness results.) Extended results for prompt robustness results. The first column is F1 and the second column is how often we capture a “null” response.}   

\end{table}   

\subsection{Few-shot prompt bias \label{results:few_shot}}    

Tables \ref{table:extended_prompt_bias_1} and \ref{table:extended_prompt_bias_2} refer to the first and second halves of the extended few-shot prompt bias task results including 95\% confidence intervals produced using 1000 bootstrap resamples, where we draw with replacement a sample that matches the size of original sample. We did not compute bootstrap confidence intervals for bias as this metric relies on balanced datasets which cannot be guaranteed with resampling.

\begin{table}[H]  

\begin{tabular}{c|cc|cc}

& \multicolumn{2}{c}{\textbf{GPT-4}} & \multicolumn{2}{c}{\textbf{GPT-3.5}}  \\\hline

\textbf{NNNN} & 0.884 [0.86, 0.91]    & 0.473     & 0.816 [0.78, 0.84]     & 0.601     	        \\    

\textbf{NNNY} & 0.887 [0.86, 0.91]    & 0.513     & 0.83 [0.8, 0.86]     & 0.569     		         \\    

\textbf{NNYY} & 0.903 [0.88, 0.92]     & 0.516     & 0.817 [0.78, 0.84]    & 0.574     	       \\  

\textbf{NYNY} & 0.9 [0.87, 0.92]    & 0.516     & 0.828 [0.8, 0.85]   & 0.561     		      \\  

\textbf{YNYN} & 0.893 [0.86, 0.91]    & 0.545     & 0.835 [0.8, 0.86]   & 0.575     		     \\  

\textbf{YYNN} & 0.895 [0.87, 0.92]   & 0.547     & 0.833 [0.8, 0.86]   & 0.578     		        \\  

\textbf{YYYN} & 0.897 [0.87, 0.92]    & 0.553     & 0.833 [0.8, 0.86]   & 0.559     		     \\  

\textbf{YYYY} & 0.883 [0.86, 0.9]    & 0.565     & 0.822 [0.79, 0.85]    & 0.572     	       \\

\end{tabular}

\caption{\label{table:extended_prompt_bias_1}(First half of extended few-shot prompt bias results.) Extended results for 4-shot bias task. The first column is F1 and the second is bias for Yes. }

\end{table}

\begin{table}[H] 

\begin{tabular}{c|cc|cc}

& \multicolumn{2}{c}{\textbf{Llama-2-7b-chat}} & \multicolumn{2}{c}{\textbf{Mistral-Instruct-01}} \\\hline

\textbf{NNNN}    & 0.483  [0.45, 0.52]      & 0.46        & 0.627 [0.59, 0.67]        & 0.185           \\    

\textbf{NNNY}    & 0.479 [0.43, 0.51]      & 0.341        & 0.551 [0.51, 0.59]        & 0.126           \\    

\textbf{NNYY}    & 0.436 [0.39, 0.47]      & 0.232        & 0.57 [0.52, 0.62]        & 0.145           \\  

\textbf{NYNY}    & 0.449 [0.4, 0.49]     & 0.224        & 0.607 [0.56, 0.65]        & 0.179           \\  

\textbf{YNYN}    & 0.341 [0.3, 0.39]     & 0.987        & 0.802 [0.77, 0.83]         & 0.45           \\  

\textbf{YYNN}    & 0.348 [0.31, 0.39]     & 0.984        & 0.804 [0.77, 0.83]        & 0.467           \\  

\textbf{YYYN}    & 0.339 [0.3, 0.38]      & 0.99        & 0.791 [0.76, 0.82]        & 0.477           \\  

\textbf{YYYY}    & 0.347 [0.31, 0.39]     & 0.984        & 0.801 [0.77, 0.83]       & 0.426           \\

\end{tabular}

\caption{\label{table:extended_prompt_bias_2}(Second half of extended few-shot prompt bias results.) Extended results for 4-shot bias task. The first column is F1 and the second is bias for Yes. }

\end{table}


%


%


%


%


%


%


%


%

%


%


\section{Appendix C – Dataset Description \label{datasets:description}}   

\subsection{PubMedQA-Labelled (PMQA-L) \label{datasets:pmqal} \cite{jin2019pubmedqa}}   

Every entry of the dataset has a question, context, a short answer and a long answer. Each entry of the dataset is derived from a PubMed article which has a question mark in the title and a structured abstract with a conclusion. The title now becomes the `question`, the abstract without the conclusive part becomes the `context` and the conclusive part becomes the `long answer`. Two annotators then annotated all the entries with whether they answer the question (yes/no/maybe).     

For our shortform evaluation tasks, we use the subset of yes/no questions and instruct the model to reply with yes or no exclusively, unless specified otherwise in the task definition. Where LLMs did not follow instructions exactly, we used RegEx to extract the first occurrence of either of the expected words. If the response did not conform with the set of expected outputs, then the response was marked as “null”.    

\subsection{BIOASQ \cite{tsatsaronis2015overview}}   

We use the dataset constructed for the BIOASQ challenge task b. These datasets are in English and contain questions, answers and relevant context. This dataset is used for two freeform evaluation tasks. The first one is a freeform QA task and in the second one we provide the model with an answer and ask it to generate a question.    

\subsection{MRPC (Microsoft Research Paraphrase Corpus) \cite{dolan2005automatically}}   

The dataset consists of sentence pairs in English. These were then hand-labelled based on being a paraphrased pair or not. This dataset is used for the text-to-text evaluation task.    

\subsection{SICK (Sentences Involving Compositional Knowledge) \label{datasets:sick} \cite{marelli-etal-2014-sick}}   
 
The dataset consists of sentence pairs in English. Through crowdsourcing efforts the dataset has been annotated for: 1) a 5-point relatedness score and 2) entailment classification with the three options being entailment, contradiction and neutral. This dataset is used for the text-to-text evaluation task. We only use the entailment classification and transform it to a binary classification task by combining contradictory and neutral pairs to “not similar”.    

\section{Appendix D – Freeform evaluation \label{section:freeform_evaluation}}   

In most real-world applications LLMs will be prompted to produce long freeform responses, the factuality and quality of which are much more challenging to evaluate. Most freeform evaluation approaches are either ground-truth based or preference-based: ground-truth based approaches compare generated text to a ground truth text \cite{BEGINbenchmark}, whereas preference-based approaches rank generated text according to criteria such as factuality, grammar or conciseness. Human judgement is often considered the gold standard for preference-based evaluations, however manual annotation can be prohibitively expensive for independent researchers, is subjective, may not scale with increasingly large evaluation datasets, and raises ethical concerns about the well-being of annotators [REF!]. Recent work demonstrates that LLMs perform in reasonable alignment with humans\cite{zheng2023judging, alpaca_eval,chiang2023can}, though LLMs can show biases when used as evaluators \cite{koo2023benchmarking}, similar to human biases \cite{clark2021all}. Due to these shortcomings of preference-based approaches, and the presence and necessity of objective ground truth in the biomedical field, we opt for semantic similarity-based freeform evaluation.    

\subsection{Semantic similarity}    

Historically, semantic similarity has been measured using ontology graphs \cite{kulmanov2021semantic}. Metrics such as ROUGE \cite{lin-2004-rouge}, BLEU \cite{BLEU} and BERTScore \cite{zhang2019bertscore} measure syntactic similarity between sentences but may fail to capture semantic similarity in syntactically different sentences. Semantic similarity can also be measured using the distance in sentence transformer embeddings \cite{reimers-2019-sentence-bert}. In this work we combine these approaches and extend the use of models as evaluators by using them to determine the semantic similarity between a long form response and a ground truth.   

\subsection{Text-to-text similarity components}    

The objective of our evaluator models was to determine the semantic similarity of an LLM output and a ground truth reference. As such, the models must take as input textual pairs and output either a binary or continuous value.     

These suffer from the following limitations. Any results observed might be biased because the model has seen the statements during one of its training stages. Also, as these are trained models, they might have acquired their own biases during training.    

\subsubsection{LLM-based}     

We provide an LLM with two texts and provide instructions on comparing them and on responding with either a “Yes” or a “No”. See above for limitations.    

\subsubsection{Unidirectional Natural Language Inference model (UNLI)}    

NLI models are trained on the task of entailment where the goal is, given two sentences, a hypothesis and a premise, to infer whether the hypothesis follows from premise (entailment), it contradicts the premise (contradiction) or whether they are unrelated (neutral). In addition to the limitations mentioned above, the model is by definition directional while the task we are interested in is not. We use the `deberta-large-mnli' model \cite{he2021deberta}.    

\subsubsection{Bidirectional Natural Language Inference (BNLI) model}    

Built by applying the UNLI model twice, once for each direction. There are multiple ways for combining the outputs of the two passes:    

\begin{itemize}   

\item Entailment \textbf{AND} Entailment $\rightarrow$ True    
\item (Entailment  \textbf{AND} Neutral) \textbf{OR} (Neutral  \textbf{AND} Entailment) $\rightarrow$ True    
\item Average the scores of the entailment values (This will lead to a continuous output)    

\end{itemize}   

While we don’t include experiments with the following methods, we include them here for reference.    

\subsubsection{Embedding-based}    

We embed both texts using a transformer model, normalise them and then take the inner product. In addition to the limitations mentioned above, a caveat of this approach is that the output is continuous rather than categorical. Even though this could have been resolved by finding a suitable threshold we do not do that in this work.    

\subsubsection{Keyword based}    

ROUGE \cite{lin2004rouge} and BLEU  \cite{BLEU} are metrics original introduced for machine translation evaluation but can also be used for textual similarity. Both ROUGE and BLEU are sets of metrics based on the overlap of n-grams between the target and reference texts. BLEU also incorporates a brevity penalty. A limitation of these approaches is that they don’t consider meaning and wouldn’t capture synonyms.    

\subsection{Evaluator performance results}    

\begin{table}[!h]   

\begin{tabular}{c|cccc}   

\hline

\textbf{}    & \textbf{LLM component} & \textbf{\begin{tabular}[c]{@{}c@{}}NLI Bidirectional \\ (Strict)\end{tabular}} & \textbf{\begin{tabular}[c]{@{}c@{}}NLI Bidirectional \\ (Relaxed)\end{tabular}} \\ \hline

MRPC - Val   & 0.661                  & 0.204                                                                          & 0.224                                                                           \\

MRPC - Test  & 0.664                  & 0.214                                                                          & 0.25                                                                            \\

SICK* - Val  & 0.798                  & 0.3                                                                            & 0.239                                                                           \\

SICK* - Test & 0.812                  & 0.294                                                                          & 0.243                                                                           \\

\end{tabular}

\caption{The metric reported here is F1. Please go to Appendix~\ref{datasets:sick} to see how the targets in SICK were modified.}

\end{table}

\end{document}